\documentclass[runningheads]{llncs}
\usepackage[T1]{fontenc}
\usepackage[utf8]{inputenc}
\usepackage{graphicx}

\usepackage{multirow}
\usepackage{makecell}
\usepackage{algorithm}
 \usepackage[bottom]{footmisc}

\usepackage{algpseudocode}

\usepackage{siunitx} 
\usepackage[hyphens]{url}
\usepackage[breaklinks]{hyperref}
\usepackage{cite}
\usepackage{caption}
\usepackage{subcaption}
\begin{document}
\title{\textbf{Towards More Efficient Data Valuation in Healthcare Federated Learning using Ensembling
\\}}
\newcommand{\orcid}[1]{\href{https://orcid.org/#1}{\textsuperscript{\includegraphics[scale=0.05]{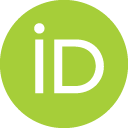}}}}
\author
{Sourav Kumar\inst{1,\dag}\orcid{0000-0003-0352-8989} \and
A. Lakshminarayanan\inst{2,\dag}\orcid{0000-0003-0429-8062} \and 
Ken Chang\inst{1}\orcid{0000-0001-6956-5059} \and
Feri Guretno\inst{2}\orcid{0000-0003-2549-4388} \and 
Ivan Ho Mien\inst{2}\orcid{0000-0002-4918-5961} \and 
Jayashree Kalpathy-Cramer\inst{1}\orcid{0000-0001-8906-9618} \and
Pavitra Krishnaswamy\inst{2,\S}\orcid{0000-0001-5893-4306} \and
Praveer Singh\inst{1,\S}\orcid{0000-0001-6641-2030}}


%
\institute{Department of Radiology, Athinoula A. Martinos Center for Biomedical Imaging, Massachusetts General Hospital, Boston, MA, USA \and
Institute for Infocomm Research, A*STAR, Singapore \\
\dag \hspace{2mm} Equally contributing first authors \\
\S \hspace{2mm} Equal senior authors \\
\email{}}
\titlerunning{Efficient Data Valuation in FL}
\authorrunning{Kumar, S. Lux, A.}
\maketitle 
\begin{abstract}

 Federated Learning (FL) wherein multiple institutions collaboratively train a machine learning model without sharing data is becoming popular. Participating institutions might not contribute equally – some contribute more data, some better quality data or some more diverse data. To fairly rank the contribution of different institutions, Shapley value (SV) has emerged as the method of choice. Exact SV computation is impossibly expensive, especially when there are hundreds of contributors. Existing SV computation techniques use approximations. However, in healthcare where the number of contributing institutions are likely not of a colossal scale, computing exact SVs is still exorbitantly expensive, but not impossible. For such settings, we propose an efficient SV computation technique called SaFE (Shapley Value for Federated Learning using Ensembling). We empirically show that SaFE computes values that are close to exact SVs, and that it performs better than current SV approximations. This is particularly relevant in medical imaging setting where widespread heterogeneity across institutions is rampant and fast accurate data valuation is required to determine the contribution of each participant in multi-institutional collaborative learning.
\keywords{Federated Learning, Data Valuation, Healthcare AI}
\end{abstract}

\section{Introduction}

Federated Learning (FL) allows machine learning (ML) models to be trained on data from multiple data contributors without the need to bring data to a central location~\cite{Li_2021}. With the growing adoption of FL in enterprise including healthcare~\cite{Rieke_2020}, it is important to quantitatively determine the contribution of individual data sources (henceforth referred to as institutions) to the performance of the global model. This data valuation technique must be fair and accurate. Shapley value (SV)~\cite{P-295}, a technique from co-operative game theory can be used to evaluate the contribution that each institution’s data bring to a global model. In a cooperative game, the contribution of each player is determined by calculating the average of all marginal contributions that the particular player brings to all possible coalitions not involving that player. SV can be used to value the data contributions of different institutions participating in FL. However, calculating SV is computationally very expensive. To determine the contribution of each institution requires training an exponential number of FL models ($2^n$ models~\cite{https://doi.org/10.48550/arxiv.1908.08619}), which becomes utterly infeasible due to astronomical computational and communication costs, even with a small number of participants. It is impossible to compute exact SV in a reasonable time when contributing institutions number in hundreds, with current techniques using approximations~\cite{FATIMA20081673,article_tmc}.

In healthcare settings, the number of contributing institutions is unlikely to be in the hundreds. For example, recent FL training research for predicting clinical outcomes of COVID-19 patients involved twenty contributing institutions~\cite{article1}. Given infrastructure and legal constraints, it is reasonable to expect 30 or fewer participating institutions in healthcare FL. Though calculating exact SV with thirty institutions is still exorbitantly expensive ($2^{30}$ FL models to compute), yet such calculations are not infeasible.

Generalizable AI models will immensely benefit from diverse data from varied sources.  This is especially relevant in healthcare because there is so much heterogeneity across medical cohorts - variability across geographies, across socio-economic levels, across different data acquisition devices and techniques. Unfortunately, healthcare data is hard to share, due to legal and ethical reasons. FL can solve this sharing problem, but contributing healthcare institutions will still want to be fairly rewarded for participating in such collaborations, specifically because annotated healthcare data - which requires specialized labelling skills, is precious. In addition, noisy labels are not uncommon in healthcare and SV computations can be useful for identifying poor quality data contributors. SV can also be used to detect malicious institutions as well as identify institutions whose contributions are marginal~\cite{DBLP:journals/corr/abs-2009-06192}.

We propose SaFE (Shapley value for Federated Learning using Ensembling) to calculate SV when number of contributing institutions is not immensely large (less than 30). SaFE uses models trained on each institution’s data as a proxy to the data itself. We still create $2^n$ models, but we use simpler models (logistic regression models), which we then aggregate using ensembling in a data-centre environment, making SaFE computationally tractable. Using empirical studies, we show that SaFE is fast, computes SV close to exact values, and performs better than existing SV approximations.

\section{Related Work}

The term Federated Learning was introduced in 2016 by Mcmahan et  al.~\cite{https://doi.org/10.48550/arxiv.1602.05629}. There are two types of FL settings - cross-device and cross-silo~\cite{https://doi.org/10.48550/arxiv.1912.04977}. Mobile device applications with thousands of devices are considered cross-device e.g. Google’s mobile keyboard prediction~\cite{https://doi.org/10.48550/arxiv.1811.03604} and Apple's “Hey Siri”~\cite{appl}, while enterprise applications where a comparatively smaller number of reliable institutions train a model (e.g. healthcare), is considered cross-silo. Cross-silo FL has been proposed in domains such as financial risk prediction~\cite{finance_ref}, drug discovery ~\cite{drug_ref,article_dr} and cybersecurity~\cite{8884802}. There is increasing interest in FL for a variety of healthcare applications e.g., image segmentation~\cite{KHOKHAR2022107818}, multi-institutional medical collaboration~\cite{articleh,Sadilek2021,Kaissis2021}, digital health~\cite{https://doi.org/10.48550/arxiv.1911.06270}, COVID-19 research ~\cite{article1} and pathology~\cite{https://doi.org/10.48550/arxiv.2009.10190}. However, data valuation for FL in healthcare applications is yet to be explored.

There has been considerable work on data valuation using SV in centralized setting, including in the medical domain~\cite{https://doi.org/10.48550/arxiv.1904.02868,https://doi.org/10.48550/arxiv.1908.08619,https://doi.org/10.48550/arxiv.1902.10275,https://doi.org/10.48550/arxiv.1911.07128,unknown}. J. Kang, Z. Xiaong et al.~\cite{8adae8295e3d4e18a60454bc3200f93d} assume prior knowledge of data quality of different participating institutions and propose mechanisms to maximize participation. SV is also used to compute feature importance for explainable AI~\cite{NIPS2017_8a20a862}. In ~\cite{unknown}, S. Tang, A. Ghorbani et al. use a technique called Data Shapley to compute the contribution of single datum to a model which is trained on a centralized chest X-Ray dataset. The computed SV is used to identify low quality data, to create better models for pneumonia detection. Since it is impossible to compute exact SV for any dataset with more than a handful of data points, Data Shapley uses Monte Carlo (MC) approximation methods, thus using only a randomly selected subset of data points in the computation. However, there are relatively limited studies on data valuation using approximate SV for FL.

The few studies exploring computation of SV for cross-device FL settings~\cite{Liu2022-ms, DBLP:journals/corr/abs-2009-06192, 9006327}, cannot consider contributions from all devices at the same time. In particular, this limits the applicability of these methods for cross-silo settings, which is relevant to medical use cases where all institutions contribute at the same time. We further note that the approximate SV computation methods may have difficulties in fairly assessing the value of data from different institutions. For example, two institutions with fairly large proportion of samples from a minority class/race/ethnicity might both be of high value compared to other institutions, but the valuation might change based on whether one or both of them have been sampled in the Monte Carlo approximation.   

\section{Background: SV Computation}
Even though it is possible to value data based on attributes such as age, volume and lineage, increasingly SV has become the method of choice for data valuation. Let $v$ denote  a utility function (performance score) with respect to which SVs are calculated. v is a mapping \(2^n\rightarrow R\), where n is the total number of players (users). The SV $\phi_i$ for the player i is defined as the average of marginal contribution the player brings to all coalitions(subsets of the players) S which do not involve the player i. So \(S=\{s\mid s \in \mathcal{P}(N) \ni i \not\in s )\}\) or we can denote ($S\subseteq N\backslash \{i\}$) where \(N=\{1,2,3,\dots,n\}\)
\begin{equation}
\phi_i(v)=\sum_{S\subseteq N\backslash \{i\} }\frac{|S|!(n-|S|-1)!}{n!}(v(S\cup \{i\})-v(S))
\end{equation}
where $|S|$ denotes the cardinality. 
Approximation methods use Monte Carlo or Truncated Monte Carlo~\cite{https://doi.org/10.48550/arxiv.1904.02868, unknown}, both of which are based on random sampling of different data permutations. This random sampling is repeated for many different permutations (until a convergence criterion is met), after which the approximate SV is computed by averaging over all calculated marginal contributions.Truncated in the truncated Monte Carlo means that the method stops parsing the current permutation and moves to a new permutation if the contribution of current is below a certain threshold. 

\section{Shapley Value for Federated Learning using Ensembling}

\textbf{Step-1: Traditional FL}: Train a FL model, using the FedAvg technique or similar model aggregation technique. At the end of FL training, every participating institution has a globally trained FL model.\\
\textbf{Step-2: Fine-Tuning}: Every participating institution uses the globally trained FL model to fine-tune a locally created model, using its own dataset. In our scheme, to enable faster SV computation in Step-3, we create a logistic regression (LR) model, trained using per-datum feature vectors extracted using a scheme similar to that proposed by S. Tang, A, Ghorbani et al.~\cite{unknown}. Unlike ~\cite{unknown}, which uses a pretrained CNN CheXNet, we use the FL model created in Step-1. This locally trained LR model is sent back to the global server.\\
\textbf{Step-3: SV using Ensembling}: On the global server, we compute all $2^n$ models using ensembling~\cite{en_ref} of LR models from Step-2. Unlike current Monte Carlo techniques to compute SV approximations, we compute SV using a simpler model (LR) and ensembling. For ensembling, we combine the Softmax predictions from each LR model to get a combined prediction. 

\begin{figure}[!htb]
    \centering
    \includegraphics[width=\textwidth]{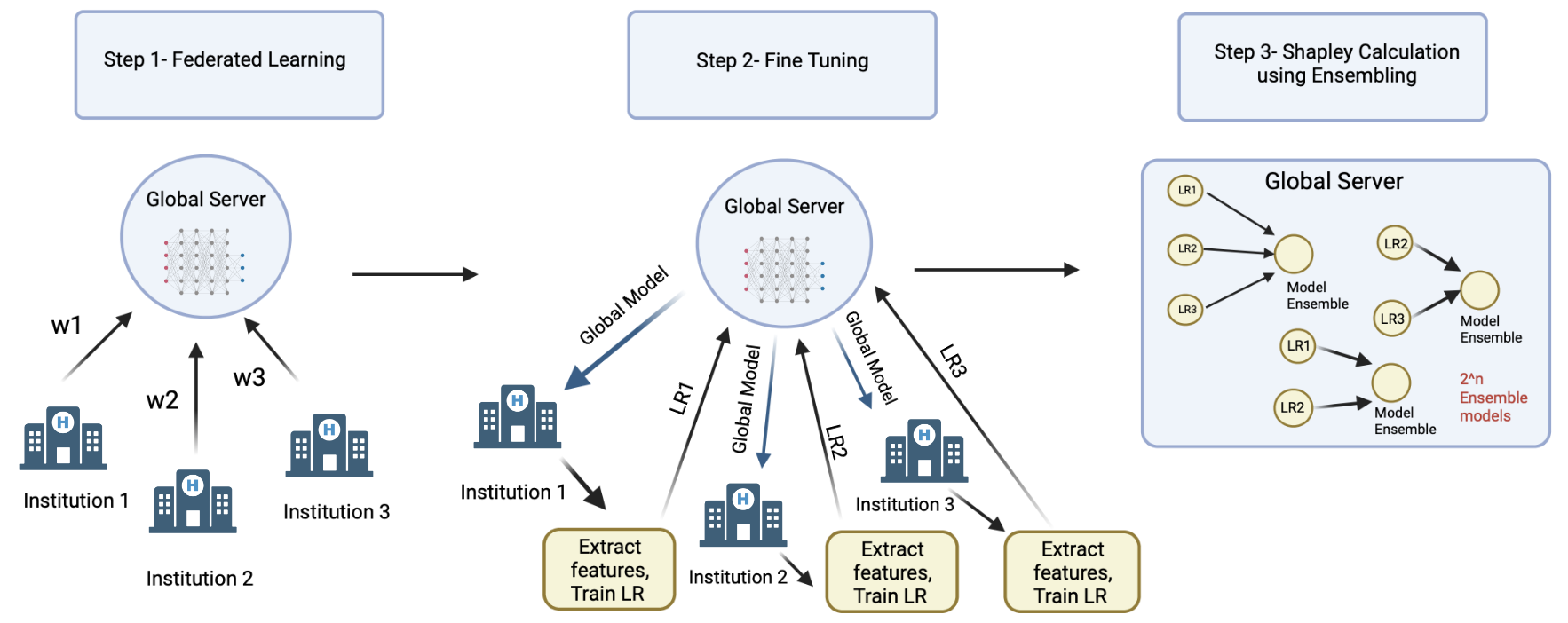}
    \caption{Shapley For Federated Learning using Ensembling (SaFE)}
    \label{fig:my_label}
\end{figure}
    \begin{algorithm}[!htb] 
\caption{Proposed Methodology to compute Shapley Value}
\label{alg:loop}
\begin{algorithmic}[1]
\Require{n participants, A global model (G), FeatureExtract- function for feature extraction from trained G, TrainLR- function to train a LR model, GetEnsemble- function that takes a subset and the set of all LR models and returns the ensemble of LR models for that subset, v- Performance score metric} 
\Ensure{$\phi_i,\dots, \phi_n$ (Shapley values for all users)}
\Statex
\State{ Initialize the global model(G) with initial weights}

\State{ Perform Federated Learning and obtain a trained Model G}
\For{each user i=1..n in parallel}
    \State{Send global model to each user}
    \State{$F_i$ $\gets$ $FeatureExtract(G,D_i)$}
    \State{$L_i$ $\gets$ $TrainLR(F_i)$}
\EndFor
\State{$\mathcal{L}\gets$$\{L_i, \dots,L_n\}$ local trained logistic regression models are made available to the central server}
\State{Initialize all $\phi_i$ $\gets$ 0}

\For{each user i=1..n }
    \State {$\mathcal{S}$ $\gets$ subsets of \{1$\dots$n\} not containing i}
    \For {s in $\mathcal{S}$}
    \State{$\phi_i$ $\gets$ $\phi_i$+ 
    $\frac{1}{n}{n-1 \choose |s|}^{-1}$}(v(GetEnsemble($\mathcal{L}$,$s$ $\cup \{i\}$)-v(GetEnsemble($\mathcal{L},s$)))

    \EndFor
\EndFor
    \State \Return {$\phi_i$,$\dots$, $\phi_n$}

\end{algorithmic}
\end{algorithm}
\textbf{Time Cost Analysis}
Training exponential number of models for even simple datasets e.g. MNIST, CIFAR takes several hours. In our experimental set-up (Linux desktop with Intel Xeon E5-2637v4 CPU, 3.5 Ghz clock, 32 GB RAM with 2 Nvidia GTX 1080 GPUs), training one FL model for MNIST (five institutions) using a simple CNN on a single GPU takes approximately 15 minutes. For a more complex dataset (ROP) and a more complex CNN, one FL model takes approximately 30 minutes. Training a 3D CNN for brain imaging using 2 GPUs takes longer, around 45 minutes. Computing exact SV for MNIST and ROP took approximately 3 and 6 hours respectively, including computing $2^5$ FL models. Note that FL training was simulated on a single machine, so data exchange between institutions and the global server was inter-process, not over the Internet, which could an order of magnitude slower and costlier.  

These time costs appear reasonable, but as the number of institutions increase, the time cost increases exponentially. For example, with 20 institutions, assuming FL training costs similar to MNIST (15 minutes per FL training), computing $2^{20}$ models will take an incredible 30 years. If we make the utterly unreasonable assumption that each institution can train 1000 models in parallel (using 1000 GPU VMs), we can reduce this time cost to 3 years. However, if there are 30 institutions, even with 1000 GPU VMs, the time-cost is 30 years! 

Using SaFE, we compute one FL model and 20 LR models (one for each of the 20 institutions) and $2^{20}$ ensembled models at the central server. Ensembling models is inexpensive. In our experimental setup creating one ensembled model takes approximately 15 milliseconds. With MNIST and ROP datasets, SaFE takes approximately 20 minutes and 35 minutes respectively. Ensembling does not require GPU and can be parallelized. With 20 institutions, we still compute $2^{20}$ models, but the time to compute SV using SaFE is 4.5 hours. For 30 institutions, if we leverage 1000 parallel cloud CPU VMs, we can still compute SV in a very reasonable 4.5 hours.

\textbf{More SaFE advantages}\\
\textbf{1.} The ensembled model created using LR models of all participating institutions performs as well as the global FL model (see result sections).\\
\textbf{2.} Ensembling LR models is much faster compared to training FL models and does not require GPU. \\
\textbf{3.} Ensembling is done on the global server, so communication costs of training $2^n$ FL trained models is not incurred. \\
\textbf{4.} SV computation (Steps 2 and 3 in Figure ~\ref{fig:my_label}) is done separate and independent of Step-1. Since Step-3 is done centrally, this allows parallelization of the SV computation.\\
\textbf{5.} Step 2 and 3 do not leak private information any more than the FL training (Step-1). Only locally trained LR model is sent to the global parameter server, no institutional data is shared with the central server.

\section{Experimental Evaluation} 
We use four different datasets - 2 well known computer vision toy datasets, namely CIFAR10 ~\cite{cifar} and MNIST~\cite{deng2012mnist}; and 2 real world medical datasets: \\
\textbf{Retinopathy Of Prematurity (ROP)}: This contains 5600 fundoscopic images from 7 different institutions classified by disease severity, namely No, Plus, and Pre-Plus~\cite{LU2022}.\\
\textbf{Brain-MRI}: Two different datasets were used to create an MRI dataset for classification. The pathological brain images are from the Brain Tumour Segmentation(BraTS) 2019 dataset~\cite{6975210} which contains 335 patients. The IXI Dataset~\cite{ixi} was used for healthy brain scans and contains 550 images. T1 contrast enhanced(T1ce) images were used from both datasets to create this dataset.

With all the 4 datasets, we perform experiments under both IID and non-IID settings. For MNIST, CIFAR10 and ROP, we split the dataset assuming 5 institutions. For brain-MRI dataset, we split assuming 4 institutions (to reduce computational costs for our experiments). We also assume that the testset is located at global server, against which the performance of different models are tested.  For each of the 4 datasets, we first extract this global testset (20$\%$ datapoints from each dataset), before creating the institutional splits. For MNIST and CIFAR the global testset is the datasets' original test split. \\
For \textbf{IID}, we use the same number of samples with nearly identical label distribution. 
Next, for \textbf{non-IID (MNIST and CIFAR10)}, we introduce non-iidness in label distribution, by keeping the number of data samples in each institution fixed but creating highly skewed label distribution splits. Institution 1 has surplus of classes 0, 1 (95$\%$ for 0,1 and rest 5$\%$ for other classes); institution 2 has surplus of classes 2,3 and so forth. 
Lastly, for \textbf{non-IID (ROP and Brain MRI)}, since both datasets have just three classes,  we maintain uneven label distribution by identifying the majority label within the two datasets and dividing that majority label amongst the different institutions in a linearly increasing fashion.
The opposite is done with the minority labels. They are distributed among the splits in a linearly decreasing fashion. This ensures an increasing percentage of majority labels and decreasing one for minority labels across splits.

Note that we chose these data splits to demonstrate the effectiveness of our SaFE method to compute SV, not to demonstrate the effectiveness of the ML learning algorithm.

\textbf{Experimental setup}
To compute exact SV in FL training, we need $2^n$ models. In out MNIST and CIFAR-10 experiments, we train $2^5$ FL models, while for Brain-MRI experiments, we train $2^4$ models. 

For ROP classification, we use a Resnet18 model with pretrained ImageNet weights. For MNIST and CIFAR10 classification, we use a simpler CNN from~\cite{https://doi.org/10.48550/arxiv.1602.05629}. For Brain-MRI, we use a 3D Resnet18 mixed convolution network~\cite{Chatterjee_2022} with pretrained weights. The learning rate for Brain-MRI classification is 2.5e-7. For CIFAR10, MNIST and ROP experiments, learning rate is 1e-4. The batch size was 64 for CIFAR10, MNIST, ROP whereas for Brain-MRI experiments, a batch size of 8 was used. We used epoch size of 20 for MNIST, ROP and Brain-MRI dataset and for CIFAR10 it was 50. For logistic regression models, the solver used was saga with elasticnet regularization with l1 ratio of 0.5. \\ 
We used a Linux desktop with Intel Xeon E5-2637v4 CPU, clocked at 3.5 GHZ, with 32 GB RAM and two Nvidia GTX 1080 GPUs. A single GPU was used for CIFAR10, MNIST and ROP experiments and both GPUs were used for Brain-MRI classification. Code-base is available on request.

\section{Results}
\begin{figure}[!htb]

\begin{subfigure}{.5\textwidth}
  \centering
  \includegraphics[scale=0.45]{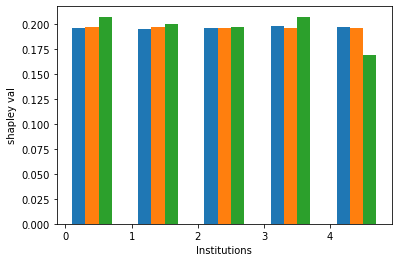}
  \label{fig:sub1}
\end{subfigure}%
\hspace{1em}
\begin{subfigure}{.5\textwidth}
  \centering
  \includegraphics[scale=0.45]{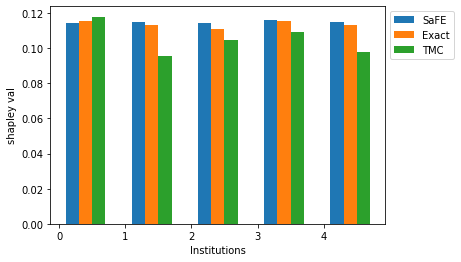}
  \label{fig:sub2}
\end{subfigure}
\caption{Shapley for MNIST-IID (left) and CIFAR-IID (right).}
\label{fig:test}
\end{figure}

\textbf{LR Ensembling vs Global FL}:
We compare the performance of the globally ensembled model created using locally fine-tuned LR models (from every contributing institution) and compare it to the performance of the FL model (Step-1 of SaFe). As shown in Table~\ref{tab:temps} the ensembled model accuracy is very similar to the performance of traditional FL model. For ROP experiments, the AUROCs are 0.96 and 0.95 for IID and non-IID splits respectively for both FL and LR-ensembled models. We observe the same with Brain-MRI as well, with both FL and LR-ensembled models having AUROCs of 0.94. 
\begin{table}[h]
\caption{Ensemble LR performance vs Traditional FL performance}
      \begin{subtable}[h]{0.25\textwidth}
        \centering
        \begin{tabular}{l| c|  c}  
      \textbf{Dataset} & \textbf{\makecell{LR Ensemble\\  Model acc}} & \textbf{\makecell{Global FL  \\ Model acc}}\\
            \hline
      MNIST & 98.28 & 98.41\\
      CIFAR10 & 82 & 83\\
      ROP & 90 & 94\\
      Brain MRI & 93 & 94\\
    \end{tabular}
        \caption{IID}
        \label{tab:week2}
     \end{subtable}
    \hspace{3cm}
    \begin{subtable}[h]{0.25\textwidth}
        \centering
        \begin{tabular}{l| c| c}  
      \textbf{Dataset} & \textbf{\makecell{LR Ensemble \\ Model acc}} & \textbf{\makecell{Global FL  \\ Model acc}}\\
            \hline
      MNIST &  96.8& 96.78\\
      CIFAR10 & 73 & 76\\
      ROP & 95 & 96\\
      Brain MRI & 89 & 91
    
    \end{tabular}
        \caption{Non-IID}
        \label{tab:week2}
     \end{subtable}
     
     \label{tab:temps}
\end{table}\\
\textbf{SV Comparison Results}:
To compare our proposed SaFE method for computing SV against exact SV computation,  we use the cosine similarity measure, commonly used to compare similarity between two vectors~\cite{cos_ref}. SV computed using SaFE are very similar to exact SV as shown in Table 2. Quite expectedly, the SV for each institution in an IID setting is almost the same. For Brain MRI and ROP datasets, we observe almost similar SV too. With non-IID splits, the similarity scores are not as close as IID splits, but still very high. 
\begin{table}[htb!]
\caption{Calculated Shapley values for different datasets using our SaFe Method}
  \begin{center}
    
    \label{tab:table1}
    \begin{tabular}{l|c|c|c|c|c|c|c|c}
      \textbf{Dataset} & \textbf{ Expt. Setting }&\textbf{ Shapley type } & \textbf{1}&\textbf{ 2 }&\textbf{ 3 }&\textbf{ 4 }&\textbf{ 5 }&\textbf{ Similarity}\\
      & & & \multicolumn{5}{c|}{Institutions}\\
      \hline
      \multirow{4}{*}{MNIST}&\multirow{2}{*}{IID}&Exact Shapley& 0.197 & 0.197 & 0.196 & 0.196 & 0.196 &\multirow{2}{*}{0.999}\\ 
      && Our Shapley &0.196& 0.197 & 0.196 & 0.197 & 0.196\\
      \cline{2-9}
      &\multirow{2}{*}{NON IID}&Exact Shapley & 0.195&0.194&0.192&0.191 &0.195 & \multirow{2}{*}{0.999}\\
      && Our Shapley & 0.186&0.197&0.199&0.190&0.193\\
      \hline
      \multirow{4}{*}{CIFAR}&\multirow{2}{*}{IID}&Exact Shapley&0.115&0.113&0.111&0.115&0.113&\multirow{2}{*}{0.999}\\
      && Our Shapley & 0.116&0.115&0.116&0.114&0.111\\
      \cline{2-9}
      &\multirow{2}{*}{NON IID}& Exact Shapley & 0.101 &0.070 &0.086 &0.115 &0.100&\multirow{2}{*}{0.973}\\
       & &Our Shapley&0.106 &0.104 &0.097 &0.080 &0.101 \\
      \hline
      \multirow{4}{*}{ROP}&\multirow{2}{*}{IID}&Exact Shapley&0.183&0.175&0.181&0.200&0.200&\multirow{2}{*}{0.997}\\
      &&Our Shapley& 0.182&0.182&0.180&0.180&0.176\\
      \cline{2-9}
      &\multirow{2}{*}{NON IID}& Exact Shapley & 0.209&0.202&0.187&0.192&0.176&\multirow{2}{*}{0.99}\\
      && Our Shapley&0.187&0.192&0.186&0.192&0.193\\
      \hline
    \multirow{4}{*}{Brain MRI}&\multirow{2}{*}{IID}&Exact Shapley&0.230&0.209&0.230&0.226&&\multirow{2}{*}{0.998}\\
    &&Our Shapley&0.218&0.216&0.209&0.230&\\
    \cline{2-9}
    &\multirow{2}{*}{NON IID}& Exact Shapley &0.235&0.222&0.230&0.225&&\multirow{2}{*}{0.99}\\
    &&Our Shapley&0.187&0.210&0.176&0.176&\\
    \hline

    
    \end{tabular}
    
  \end{center}
  
\end{table}\\
\textbf{TMC vs SaFE}:
 To compute Truncated Monte Carlo (TMC)  SV, we perform random sampling of different permutations of the LR models. We used the implementation by~\cite{https://doi.org/10.48550/arxiv.1904.02868} but adapted it to work in a FL setting.  SV is calculated by parsing through these permutations and calculating the marginal contribution of every new institution once its added to the existing list of institutions already scanned. This marginal contribution is the difference in performance of the ensemble model due to the added institution while parsing a permutation. This process is repeated for many different permutations and the final SV is the average over all calculated marginal contributions. This technique doesn't consider all possible model ensembles, since it stops sampling permutations after a threshold. As seen in Figure~\ref{fig:test}, in comparison to TMC, SaFE is much closer to exact SV.

  \section{Conclusion}

When healthcare institutions participate in collaborative FL training, the contributions that their data make to the global model might not be equal. Some institutions might contribute more data, some better quality data or some more diverse data. To fairly rank the data valuation of datasets, Shapley value (SV) has emerged as the method of choice. But SV computation is impossibly expensive, when there are immensely large number of participating institutions. Even in healthcare FL, where we have a sizeable number of participants, calculating SV can be exorbitant. Existing SV techniques use approximations, which can result in unfair SV attributions. In this paper, we propose an efficient SV technique called SaFE (Shapley value for Federated Learning using Ensembling), that relies on "model" approximation (ensembling being an instance of it) instead of "SV computation" approximation. We show empirically that SaFe computes SV faster, its SV are close to exact SV, and that SaFe performs better than current approximation techniques. Future work would deepen the theoretical foundation of SaFE to obtain guarantees for different model approximation scenarios.

\let\textbf\relax
\newpage \bibliography{bibliography}

\begin{thebibliography}{10}
\providecommand{\url}[1]{\texttt{#1}}
\providecommand{\urlprefix}{URL }
\providecommand{\doi}[1]{https://doi.org/#1}

\bibitem{appl}
{Apple WWDC 2019}, \url{https://developer.apple.com/videos/play/wwdc2019/708}

\bibitem{cos_ref}
{Cosine Similarity},
  \url{https://www.sciencedirect.com/topics/computer-science/cosine-similarity}

\bibitem{en_ref}
{Ensemble Models},
  \url{https://www.sciencedirect.com/topics/computer-science/ensemble-modeling}

\bibitem{ixi}
Ixi dataset, \url{https://brain-development.org/ixi-dataset/}

\bibitem{drug_ref}
{VentureBeat: Federated learning platform for drug discovery},
  \url{https://venturebeat.com/2020/09/17/major-pharma-companies-including-novartis-and-merck-build-federated-learning-platform-for-drug-discovery/9}

\bibitem{finance_ref}
{WeBank: WeBank and Swiss Re sign cooperation MOU, 2019,},
  \url{https://www.fedai.org/news/webank-and-swiss-re-signed-cooperation-mou/}

\bibitem{article_tmc}
Castro, J., Gomez, D., Tejada, J.: {Polynomial calculation of the Shapley value
  based on sampling}. Computers $\&$ Operations Research  \textbf{{36}},
  1726--1730 (05 2009)

\bibitem{Chatterjee_2022}
Chatterjee, S., Nizamani, F.A., Nürnberger, A., Speck, O.: {Classification of
  brain tumours in {MR} images using deep spatio spatial models}. Scientific
  Reports  \textbf{12}(1) (jan 2022)

\bibitem{article_dr}
Choudhury, O., Park, Y., Salonidis, T., Gkoulalas-Divanis: {Predicting Adverse
  Drug Reactions on Distributed Health Data using Federated Learning}. AMIA
  Symposium  \textbf{2019},  313--322 (03 2020)

\bibitem{article1}
Dayan, I., Roth, H., Zhong, A., Harouni: {Federated learning for predicting
  clinical outcomes in patients with COVID-19}. Nature Medicine  \textbf{27},
  ~1--9 (10 2021)

\bibitem{deng2012mnist}
Deng, L.: {The MNIST database of handwritten digit images for machine learning
  research}. IEEE Signal Processing Magazine  \textbf{29}(6),  141--142 (2012)

\bibitem{FATIMA20081673}
Fatima, S.S., Wooldridge, M., Jennings, N.R.: {A linear approximation method
  for the Shapley value}. Artificial Intelligence  \textbf{172}(14),
  1673--1699 (2008)

\bibitem{https://doi.org/10.48550/arxiv.1904.02868}
Ghorbani, A., Zou, J.: Data shapley: Equitable valuation of data for machine
  learning. In: Proceedings of the 36th International Conference on Machine
  Learning. vol.~97, pp. 2242--2251. PMLR, Long Beach, CA (09--15 Jun 2019)

\bibitem{https://doi.org/10.48550/arxiv.1811.03604}
Hard, A., Rao, K., Mathews, R., Beaufays, F.: {Federated Learning for Mobile
  Keyboard Prediction}. ArXiv  \textbf{abs/1811.03604} (2018)

\bibitem{https://doi.org/10.48550/arxiv.1902.10275}
Jia, R., Dao, D., Wang, B., Hubis, F.A., Hynes, N., G{\"u}rel, N.M., Li, B.,
  Zhang, C., Song, D.X., Spanos, C.J.: {Towards Efficient Data Valuation Based
  on the Shapley Value}. In: AISTATS. Naha, Okinawa, Japan (2019)

\bibitem{https://doi.org/10.48550/arxiv.1908.08619}
Jia, R., Dao, D., Wang, B., Hubis, F.A., Gurel, N.M., Li, B., Zhang, C.,
  Spanos, C., Song, D.: {Efficient Task-Specific Data Valuation for Nearest
  Neighbor Algorithms}. Proc. VLDB Endow.  \textbf{12}(11),  1610–1623 (Jul
  2019)

\bibitem{https://doi.org/10.48550/arxiv.1911.07128}
Jia, R., Wu, F., Sun, X., Xu, J., Dao, D., Kailkhura, B., Zhang, C., Li, B.,
  Song, D.: {Scalability vs. Utility: Do We Have to Sacrifice One for the Other
  in Data Importance Quantification?} In: 2021 IEEE/CVF Conference on Computer
  Vision and Pattern Recognition (CVPR). pp. 8235--8243. Nashville, TN, USA
  (2021)

\bibitem{https://doi.org/10.48550/arxiv.1912.04977}
Kairouz, P., McMahan, H.B., Avent, B.: {Advances and Open Problems in Federated
  Learning}. Foundations and Trends in Machine Learning  \textbf{14}(1–2),
  1--210 (2021)

\bibitem{Kaissis2021}
Kaissis, G., Ziller, A., Passerat-Palmbach, J., Ryffel, T., Usynin, D.:
  {End-to-end privacy preserving deep learning on multi-institutional medical
  imaging}. Nature Machine Intelligence  \textbf{3}(6),  473--484 (May 2021)

\bibitem{8adae8295e3d4e18a60454bc3200f93d}
Kang, J., Xiong, Z.: {Incentive Mechanism for Reliable Federated Learning: A
  Joint Optimization Approach to Combining Reputation and Contract Theory}.
  IEEE Internet of Things Journal  \textbf{6}(6),  10700--10714 (Dec 2019)

\bibitem{KHOKHAR2022107818}
KhoKhar, F.A., Shah, J.H.: {A review on federated learning towards image
  processing}. Computers and Electrical Engineering  \textbf{99},  107818
  (2022)

\bibitem{cifar}
Krizhevsky, A., Nair, V., Hinton, G.: Cifar-10
  \url{http://www.cs.toronto.edu/~kriz/cifar.html}

\bibitem{Li_2021}
Li, Q., Wen, Z.: {A Survey on Federated Learning Systems: Vision, Hype and
  Reality for Data Privacy and Protection}. {IEEE} Transactions on Knowledge
  and Data Engineering pp.~1--1 (2021)

\bibitem{Liu2022-ms}
Liu, Z., Chen, Y., Yu, H., Liu, Y., Cui, L.: {GTG-Shapley}: Efficient and
  accurate participant contribution evaluation in federated learning. ACM
  Trans. Intell. Syst. Technol.  \textbf{13}(4),  1--21 (Aug 2022)

\bibitem{LU2022}
Lu, C., Hanif, A., Singh, P., Chang, K., Coyner, A.S., Brown, J.M., Ostmo, S.,
  Chan, R.V.P., Rubin, D., Chiang, M.F., Campbell, J.P., Kalpathy-Cramer, J.:
  {Federated Learning for Multi-center Collaboration in Ophthalmology:
  Improving Classification Performance in Retinopathy of Prematurity}.
  Ophthalmology Retina  (2022)

\bibitem{https://doi.org/10.48550/arxiv.2009.10190}
Lu, M.Y., Chen, R.J., Kong, D., Lipkova, J., Singh, R., Williamson, D.F., Chen,
  T.Y., Mahmood, F.: {Federated learning for computational pathology on
  gigapixel whole slide images}. Medical Image Analysis  \textbf{76},  102298
  (2022)

\bibitem{NIPS2017_8a20a862}
Lundberg, S.M., Lee, S.I.: A unified approach to interpreting model
  predictions. In: Guyon, I., Luxburg, U.V., Bengio, S., Wallach, H., Fergus,
  R., Vishwanathan, S., Garnett, R. (eds.) Advances in Neural Information
  Processing Systems. vol.~30. Curran Associates, Inc., Long Beach,CA (2017)

\bibitem{https://doi.org/10.48550/arxiv.1602.05629}
McMahan, B., Moore, E., Ramage: {Communication-Efficient Learning of Deep
  Networks from Decentralized Data}. In: Proceedings of the 20th International
  Conference on Artificial Intelligence and Statistics. Proceedings of Machine
  Learning Research, vol.~54, pp. 1273--1282. PMLR, Ft. Lauderdale, FL (20--22
  Apr 2017)

\bibitem{6975210}
Menze, B.H., Jakab: {The Multimodal Brain Tumor Image Segmentation Benchmark
  (BRATS)}. IEEE Transactions on Medical Imaging  \textbf{34}(10),  1993--2024
  (2015)

\bibitem{8884802}
Nguyen, T.D., Marchal, S., Miettinen, M., Fereidooni: {DÏoT: A Federated
  Self-learning Anomaly Detection System for IoT}. In: 2019 IEEE 39th
  International Conference on Distributed Computing Systems (ICDCS). pp.
  756--767. Dallas, Texas (2019)

\bibitem{Rieke_2020}
Rieke, N., Hancox, J., Li, W., Milletar{\`{\i}}, F., Roth, H.R., Albarqouni,
  S.: {The future of digital health with federated learning}. npj Digital
  Medicine  \textbf{3}(1) (sep 2020)

\bibitem{Sadilek2021}
Sadilek, A., Liu, L., Nguyen, D., Kamruzzaman, M., Serghio, S.: {Privacy-first
  health research with federated learning}. npj Digital Medicine  \textbf{4}(1)
  (Sep 2021)

\bibitem{P-295}
Shapley, L.S.: A Value for N-Person Games. RAND Corporation, Santa Monica, CA
  (1952)

\bibitem{articleh}
Sheller, M., Edwards, B., Reina, G., Martin, J., Pati, S., Kotrotsou:
  {Federated learning in medicine: facilitating multi-institutional
  collaborations without sharing patient data}. Scientific Reports  \textbf{10}
  (07 2020)

\bibitem{9006327}
Song, T.: Profit allocation for federated learning. In: 2019 IEEE International
  Conference on Big Data (Big Data). pp. 2577--2586 (2019).
  \doi{10.1109/BigData47090.2019.9006327}

\bibitem{unknown}
Tang, S., Ghorbani, A., Yamashita, R., Rehman, S., Dunnmon, J.A., Zou, J.,
  Rubin, D.L.: {Data valuation for medical imaging using Shapley value and
  application to a large-scale chest X-ray dataset}. Sci. Rep.  \textbf{11}(1),
  ~8366 (Apr 2021)

\bibitem{DBLP:journals/corr/abs-2009-06192}
Wang, T., Rausch, J., Zhang, C., Jia, R., Song, D.: {A Principled Approach to
  Data Valuation for Federated Learning}. CoRR  \textbf{abs/2009.06192} (2020)

\bibitem{https://doi.org/10.48550/arxiv.1911.06270}
Xu, J., Glicksberg, B.S., Su, C., Walker, P., Bian, J., Wang, F.: {Federated
  learning for healthcare informatics}. J. Healthc. Inform. Res.
  \textbf{5}(1),  1--19 (2021)

\end{thebibliography}
\end{document}